\documentclass[letterpaper]{article} % DO NOT CHANGE THIS
\usepackage[submission]{aaai23}  % DO NOT CHANGE THIS
\usepackage{times}  % DO NOT CHANGE THIS
\usepackage{helvet}  % DO NOT CHANGE THIS
\usepackage{courier}  % DO NOT CHANGE THIS
\usepackage[hyphens]{url}  % DO NOT CHANGE THIS
\usepackage{graphicx} % DO NOT CHANGE THIS
\usepackage{natbib}  % DO NOT CHANGE THIS AND DO NOT ADD ANY OPTIONS TO IT
\usepackage{caption} % DO NOT CHANGE THIS AND DO NOT ADD ANY OPTIONS TO IT
\usepackage{algorithm}
\usepackage{algorithmic}
\usepackage{graphicx, amsmath, bbm}
\usepackage{multirow}
\usepackage{enumitem}
\usepackage[normalem]{ulem}
\useunder{\uline}{\ul}{}
\usepackage{makecell}
\usepackage{amssymb}
\usepackage{booktabs}
\usepackage{color}

\usepackage{newfloat}
\usepackage{listings}
\usepackage[colorlinks, linkcolor=red]{hyperref}
\usepackage[misc]{ifsym} 
\DeclareCaptionStyle{ruled}{labelfont=normalfont,labelsep=colon,strut=off} % DO NOT CHANGE THIS
\floatstyle{ruled}
\newfloat{listing}{tb}{lst}{}
\floatname{listing}{Listing}

\title{Adaptive Perception Transformer for Temporal Action Localization}
\author{
    Yizheng Ouyang$^1$,
    Tianjin Zhang$^2$,
    Weibo Gu$^3$,
    and Hongfa Wang$^4$
}
\affiliations{

Tianjin University$^1$\thanks{oyyizheng@tju.edu.cn} \quad Sichuan University$^2$ \quad Huazhong University of Science and Technology$^3$ \\ Chinese Academy of Sciences$^4$

}

\usepackage{bibentry}

\begin{document}

\maketitle

\begin{abstract}
Temporal action localization aims to predict the boundary and  category of each action instance in untrimmed long videos. Most of previous methods based on anchors or proposals neglect the global-local context interaction in entire video sequences. Besides, their multi-stage designs cannot generate action boundaries and categories straightforwardly. To address the above issues, this paper proposes a end-to-end model, called Adaptive Perception transformer (AdaPerFormer for short). Specifically, AdaPerFormer explores a dual-branch attention mechanism. One branch takes care of the global perception attention, which can model entire video sequences and aggregate global relevant contexts. While the other branch concentrates on the local convolutional shift to aggregate intra-frame and inter-frame information through our bidirectional shift operation. The end-to-end nature produces the boundaries and categories of video actions without extra steps. Extensive experiments together with ablation studies are provided to reveal the effectiveness of our design. Our method obtains competitive performance on the THUMOS14 and ActivityNet-1.3 dataset.
\end{abstract}

\section{Introduction}
Temporal action localization (TAL) is a challenging problem in video understanding, which aims to predict action categories and regions contained in untrimmed videos. 
The main difficulty of TAL lies in the frame-level prediction of ambiguous temporal boundaries.
Furthermore, the durations of different actions in TAL varies from a few seconds to hundreds of seconds, which makes accurate action classification difficult.
\begin{figure}[t]
\centering
\includegraphics[width=1\linewidth]{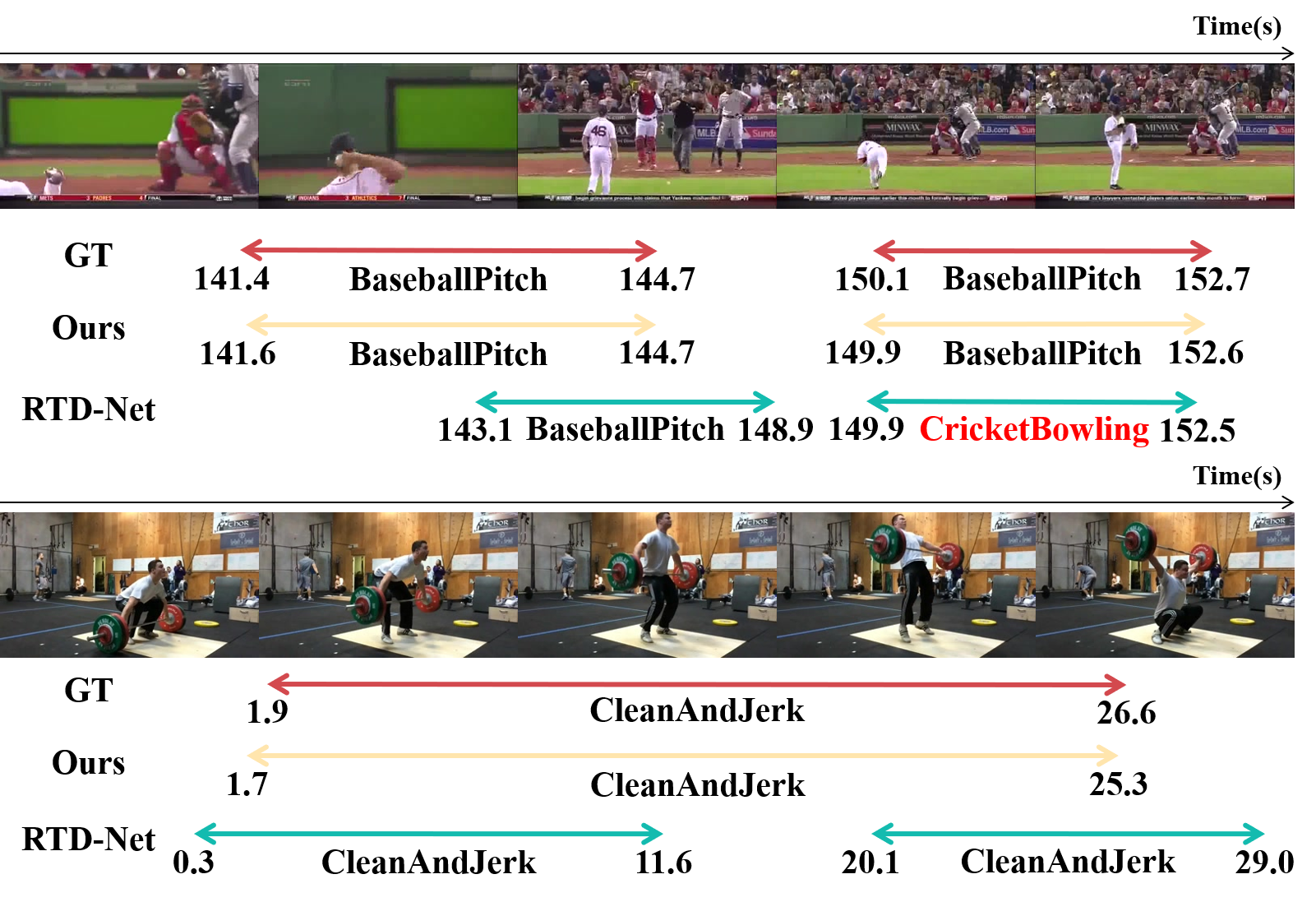}
\caption{Two examples of TAL. The query-based method RTD-Net \cite{DBLP:conf/iccv/TanT0W21} produces inaccurate boundaries/categories or separated temporal regions, while our AdaPerFormer generates accurate results in an end-to-end mode. }
\label{visual}
\end{figure}
To handle these issues, numerous anchor-based \cite{DBLP:conf/cvpr/ChaoVSRDS18,DBLP:conf/aaai/LiuW20,DBLP:conf/cvpr/Lin0LWTWLHF21} and proposal-based \cite{DBLP:conf/iccv/ZhaoXWWTL17,DBLP:conf/iccv/LinLLDW19,DBLP:conf/aaai/LinLWTLCWLHJ20,DBLP:conf/cvpr/QingSGW0W0YGS21} methods have been proposed. Recent deep models like  \cite{DBLP:conf/iccv/TanT0W21, DBLP:journals/corr/abs-2106-10271} rely on the self-attention operations.
Though these methods have made steady progress, three key limitations remain:
\textbf{First}, the methods based on anchors or proposals are susceptible to hand-crafte anchors and confidence thresholds.
\textbf{Second}, the existing methods cannot adaptively handle the various action scenes in videos. Due to the wide distribution of video lengths and the various action durations in a video, some strategies adopt slicing video sequences or generating numerous proposals. However, these modes would break the semantic information of complete feature sequences and make it challenging to acquire global and local contexts in the entire video sequences. As shown in Fig. \ref{visual}, existing methods cannot effectively estimate the complete segments for long actions.
\textbf{Third}, the primitive self-attention mechanism neglects the continuity of each video action. This attention mechanism needs to calculate the correlations of all positions, which ignores the temporal distance and order of video sequences. Besides, dense attention suffers from high computation costs and slow convergence. Since the local context during an action is more informative than the context at a distance, existing approaches cannot utilize these local contexts to classify continuous actions.

In this paper, we propose a new temporal action localization framework based on Transformer, named AdaPerFormer. 
More concretely, we customize an temporal-aware attention (TAA) mechanism, which enables our model to directly process the entire feature sequences in an end-to-end manner. Under the premise of guaranteeing the minimum calculation cost, our method achieves the local interaction of temporal context information while modeling long-range global dependency. Our TAA contains two branches, including the global perception attention and the local convolution shift. To preserve the continuity of the action regions, the global perception attention branch presents a low-cost self-attention method, which constructs the interaction of each relevant position in the complete sequence. 
The local convolutional shift branch models the local context of adjacent frame features. It entirely connects the local interaction between intra-frame and inter-frame information by performing shift operations along temporal and channel dimensions. Moreover, our simple encoder-decoder design avoids the complex process of placing anchors or generating temporal proposals, and can directly predict action boundaries and categories without extra manipulation.

Our proposed AdaPerFormer exhibits promising capability in TAL and achieves the state-of-the-art performance on the standard benchmarks. Specifically, AdaPerFormer obtains an average mAP 59.3\% on the THUMOS14 dataset. This accuracy outperforms the current methods based on anchors or proposals. Furthermore, AdaPerFormer achieves competitive results on the ActivityNet-1.3 with an average mAP of 35.1\%, getting higher performance than most previous methods. The major contributions of this work are summarized as follows:
\begin{itemize}
\item We propose a end-to-end temporal action localization framework named AdaPerFormer, which can generate action categories and boundaries directly without artificial anchors or proposals.
\item We propose an temporal-aware attention mechanism. The dual-branch architecture flexibly models global contexts in entire video sequences and effectively realizes local interaction between intra-frame and inter-frame information.
\item Our method obtains superior results than other state-of-the-art alternatives on the THUMOS14 and achieves competitive performance on the ActivityNet-1.3.
\end{itemize}

\section{Related Work}
Existing temporal action localization methods can be roughly categorized into three groups as follows.
\subsection{Anchor-based methods}
These methods \cite{DBLP:conf/mm/LinZS17,DBLP:conf/iccv/XuDS17,DBLP:conf/iccv/GaoYSCN17,DBLP:conf/cvpr/ChaoVSRDS18,DBLP:conf/aaai/LiuW20,DBLP:conf/cvpr/Lin0LWTWLHF21,DBLP:conf/cvpr/LongYQTLM19} need to pre-define a series of sliding windows or anchor temporal regions of different durations first, then determine whether each of them contains actions and finetune the boundaries.
CBR  \cite{DBLP:conf/bmvc/GaoYN17} proposed a cascaded boundary regression model which can be used both in temporal proposals generating and action detecting.
PBRNet  \cite{DBLP:conf/aaai/LiuW20} designed an end-to-end refinement framework, which uses three modules to detect different scales of actions and refine the boundaries of actions.
A2Net \cite{DBLP:journals/tip/YangPZFH20} utilized an anchor-free mechanism to predict the distance from each position in the feature sequences to the temporal boundary.
ASFD \cite{DBLP:conf/cvpr/Lin0LWTWLHF21} explored a boundary consistency learning strategy that can constrain the model to learn better boundary features.
Anchor-based methods need the heuristic rules such as the size and number of anchors and lack boundary precision that is influenced by a unit minimum length. 
Differently, our method can achieve high boundary precision at frame level without anchors.
\subsection{Boundary-based methods}
Unlike anchor-based methods, boundary-based methods  \cite{DBLP:conf/eccv/LinZSWY18, DBLP:conf/iccv/LinLLDW19, DBLP:conf/eccv/ZhaoXJZW020, DBLP:conf/aaai/LinLWTLCWLHJ20,DBLP:conf/cvpr/QingSGW0W0YGS21,DBLP:conf/eccv/BaiWTYLL20,DBLP:conf/iccv/ZhuT00021, DBLP:conf/aaai/ChenZWL22} predict the start and end moments of the video action locally, then combines the start and end moments into temporal proposals, and finally performs category prediction for each proposal. 
BSN  \cite{DBLP:conf/eccv/LinZSWY18} aimed to combine local temporal boundaries and global confidence scores.
To simplify the pipeline, BMN  \cite{DBLP:conf/iccv/LinLLDW19} proposed the boundary matching mechanism, which denotes proposal as a pair and converts all proposals to  a Boundary-Matching confidence map.
DBG  \cite{DBLP:conf/aaai/LinLWTLCWLHJ20} designed a dense boundary generator to predict the boundary map.
DCAN \cite{DBLP:conf/aaai/ChenZWL22} aggregated contexts on boundary level and proposal level to generate high-quality action proposals, thereby improving the performance of temporal action detection.
These methods only generate temporal proposals and then use an additional action classifier, so they cannot jointly optimize the action and temporal precision.
In our method,  we can simultaneously predict action durations and categories in a single network by end-to-end training. 
\subsection{Transformer in TAL}
Recently, With the introduction of ViT \cite{DBLP:conf/iclr/DosovitskiyB0WZ21}, the transformer architecture has recently achieved remarkable results in vision tasks. Subsequent studies, including Swin Transformer \cite{DBLP:conf/iccv/LiuL00W0LG21}, Segformer \cite{DBLP:conf/nips/XieWYAAL21}, and Deformable DERT \cite{DBLP:conf/iclr/ZhuSLLWD21}, further advanced the widespread use of transformer in classification, segmentation, and detection tasks. In TAL, inspired by the advance of transformer, several works generated proposals by using localization queries instead of anchors.
RTD-Net  \cite{DBLP:conf/iccv/TanT0W21} utilized the transformer decoder to interact the query with the globally encoded features to generate better proposals; 
Tad-TR  \cite{DBLP:journals/corr/abs-2106-10271} extracted the temporal context information required for action prediction by selectively processing a set of sparse segments in the video. 
ActionFormer \cite{DBLP:journals/corr/abs-2202-07925} proposed a simple yet powerful model which combines the multi-scale transformer encoder and light-weighted decoder to examine each moment in time.
In our method, we explore an end-to-end transformer architecture in TAL, which aims to capture the global dependencies of temporal features while interacting with local contextual information.
\begin{figure}[t]
  \centering
  \includegraphics[width=1\linewidth]{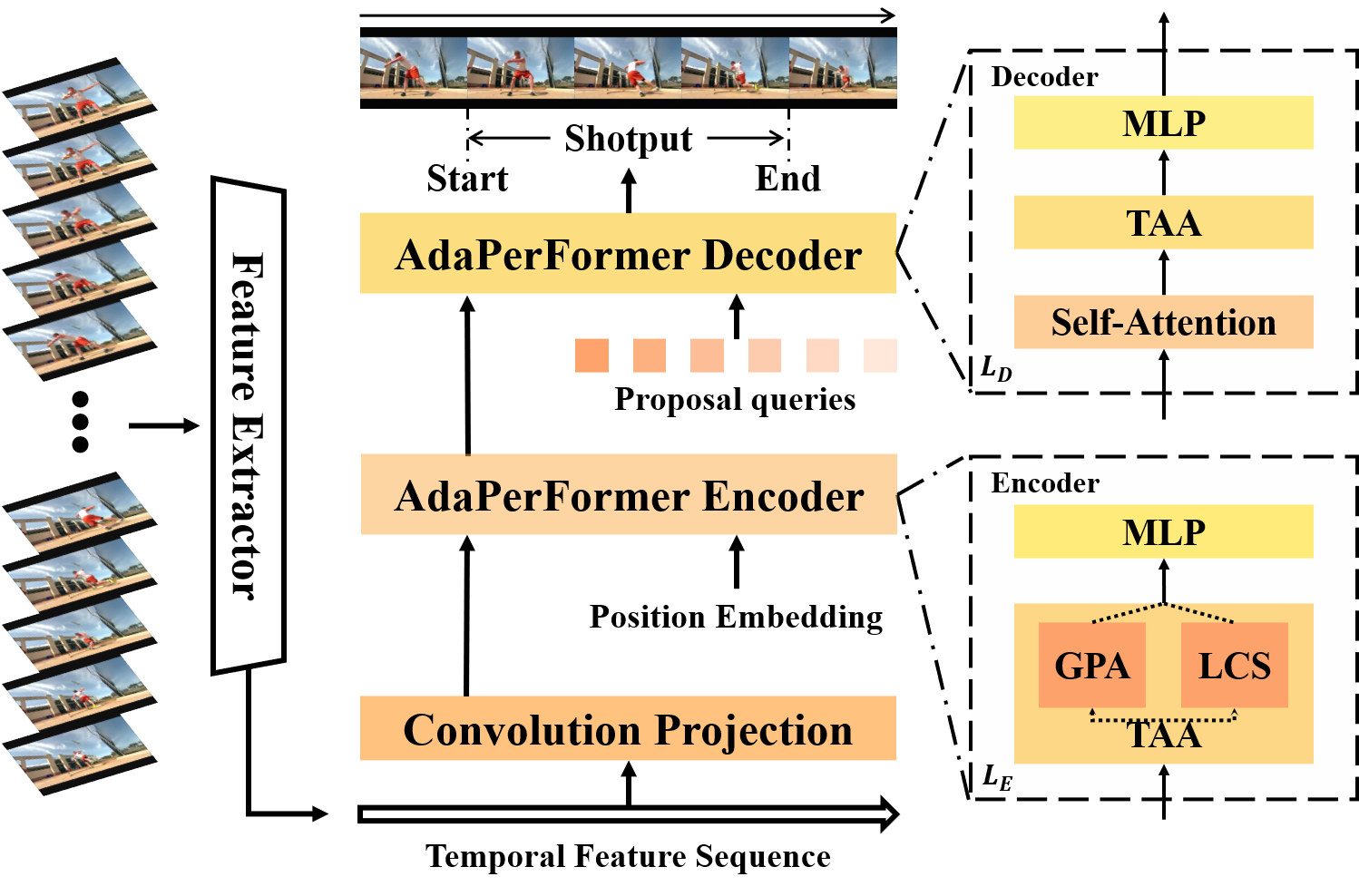}
  \caption{Overview of our AdaPerFormer. The Encoder-Decoder part consists of temporal-aware attention (TAA) and multilayer perceptron (MLP), where TAA contains two branches of global perception attention (GPA) and local convolutional shift (LCS).}
  \label{AdaPerFormer}
  \vspace{2pt}
\end{figure}
\section{Method}
AdaPerFormer is an end-to-end framework for TAL. Given an input video as a frame sequence $X = \{x_t\}^T_{t=1}$ with $T$ frames, where $x_t$ represent the $t$-th frame, the pre-trained video action recognition backbone will extract a continuous temporal feature sequence $S$. 
Our encoder-decoder architecture makes hierarchical modeling on these temporal feature sequences after convolutional projection, estimates the confidence scores of the predicted proposals, and directly generates the action categories and boundaries. We express the detection results as $\Phi_X$ = $\{(\phi_n, c_n)\}^{N_X}_{n=1}$, where $\phi_n = (t_s, t_e)$ denotes the start time and end time of the $n$-th instance, $c_n$ indicates its action label, and $N_X$ represents the number of detected action instances.
\vspace{-2pt}
\subsection{AdaPerFormer}
To obtain global contexts from the complete feature sequences and classify actions precisely, we notice the superior performance of self-attention mechanisms on vision tasks. Several previous image attention mechanisms \cite{DBLP:journals/pami/HuSASW20, DBLP:conf/nips/HuSASV18} show that self-attention can capture long-range structure of global context. Unlike those methods that use feature sequences built by image patches, the features in TAL task extracted by the action recognition backbone have natural temporality. These sequential features are more suitable for integration with the transformer-based architecture. Therefore, we build our AdaPerFormer as shown in Figure \ref{AdaPerFormer}.
Let $S_v \in \mathbb{R}^{C \times T}$ denote these sequential video features, where $C$ and $T$ are the dimension and sequence length respectively. Similar to recent research \cite{DBLP:conf/nips/LiJXZCWY19, DBLP:conf/nips/XiaoSMDDG21}, we use a lightweight convolutional network to project features into a $C_D$-dimensional space, which helps the transformer encoder aggregate local contextual information in time. Also, the position embedding will be initialized at first. Our encoder takes the features and embeddings as input $E_{in} \in \mathbb{R}^{C_D \times T}$, and outputs embeddings $E_e \in \mathbb{R}^{C_D \times T}$ after $L_E$ transformer layers. Each layer contains TAA and MLP blocks after a LayerNorm, and the residual connection is applied after both blocks. We can express the process of the $l^{th}$ layer as:
\begin{equation}
    \centering
    \begin{aligned}
    E^\mathit{l}_\mathit{h1} &= \operatorname{TAA}(\operatorname{LN}(E^\mathit{l - 1})) + E^\mathit{l - 1}, 
    \\
    E^\mathit{l}_\mathit{h2} &= \operatorname{MLP}(\operatorname{LN}(E^\mathit{l}_\mathit{h1})) + E^\mathit{l}_\mathit{h1}, 
    \end{aligned}
\end{equation}
where $E^\mathit{l - 1}, E^\mathit{l}_\mathit{h1}, E^\mathit{l}_\mathit{h2} \in \mathbb{R}^{C_D \times T}$. After encoder, the decoder takes the embeddings $E_e$ and a set of learnable proposal queries as input. Since the continuous actions generally show local correlation on the feature sequence, decoupling local information in embeddings is beneficial for action classification. Our layer-by-layer decoder structure can model the local context between embeddings, and output $E_{out}$ for prediction. 
Following previous methods\cite{DBLP:conf/iccv/TanT0W21, DBLP:journals/corr/abs-2106-10271} using a transformer decoder, we use the proposal matching method to assign the predictions to the ground truth. We use the classification head to predict sequence's action probability $c_n$ for each proposal query. While another regression head predicts the onset and offset $(t_s , t_e)$ as the action boundary.
The main strength of our model is the ability to adaptively model the global-local context dependencies in entire sequences, which is beneficial from our designed TAA mechanism.
\begin{figure*}[t]
  \centering
  \includegraphics[width=1\linewidth]{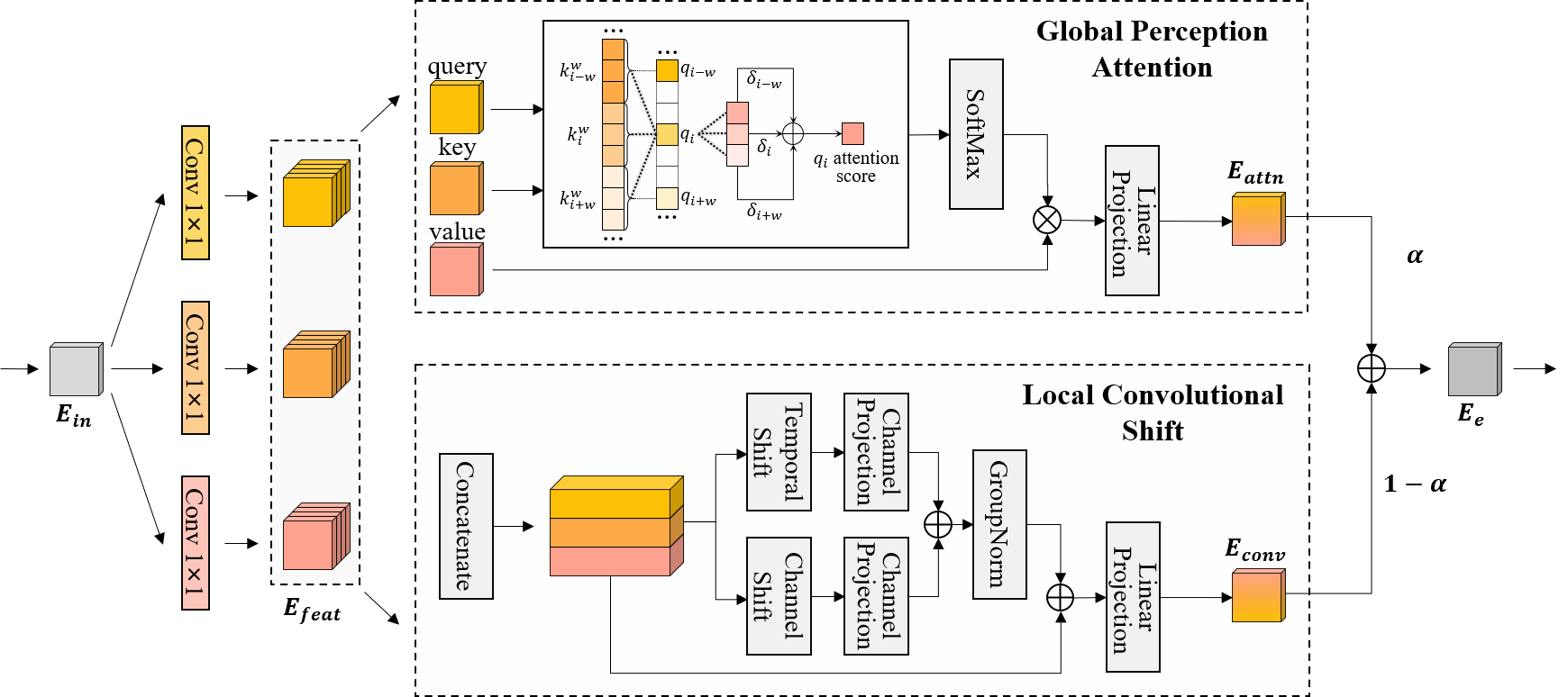}
  \caption{Overview of our TAA. Our TAA contains two branches. The global perception attention branch uses a novel window attention to construct the global representation of entire feature sequence, and the local convolutional shift branch uses bidirectional shift to better model the local interaction in frame-level. Both branches work together to generate more accurate action boundaries and categories.}
  \label{TAA}
\end{figure*}
\subsection{Temporal-aware attention}
The spatio-temporal complexity of the original self-attention mechanisms is quadratic with the sequence length, which largely limits the input of the model. Previous method \cite{DBLP:conf/iclr/DosovitskiyB0WZ21} truncate overly long sequences as input. But in TAL task, this truncation operation will destroy the temporality and completeness of the sequence. For example, in a video containing the action of `CleanAndJerk', the action of weightlifting runs through almost the entire video content. If the feature sequence of the video is truncated, the continuity of the action will be destroyed, which makes it difficult to classify the action. 
Therefore, reducing the computational complexity while maintaining the maximum time resolution of the sequence is necessary.
Moreover, the onset and offset of action are often within a small range of a few frames. To generate more accurate action boundaries, we are supposed to fully exploit the local contexts between adjacent frames with a local convolution operation. 
To combine self-attention and convolution efficiently, inspired by previous works \cite{DBLP:conf/cvpr/HuSS18, DBLP:conf/iccv/WuXCLDY021}, we decompose the convolution operation into the following two steps:
\begin{equation}
\centering
\begin{aligned}
&\mathbf{I}: \tilde{l}_{i j}^{(m, n)}=K_{m, n} s_{i j} , \\
&\mathbf{II}:l_{i j}=\sum_{m, n} l_{i j}^{(m, n)},
\\
&=\sum_{m, n}{\operatorname{Shift}\left(\tilde{l}_{i j}^{(m, n)}, m-\lfloor k / 2\rfloor, n-\lfloor k / 2\rfloor\right) ,}\hfill
\end{aligned}
\label{shiftop}
\end{equation}
where $s_{ij}, l_{ij}$ are the feature tensors of pixel$(i, j)$, $K_{m, n}$ represents the kernel weights with regard to the indices of the kernel position $(m, n)$, and $k$ represents kernel size. The shift operations $\tilde{f}: \operatorname{Shift}(f, \Delta x, \Delta y)$ in Eq. \eqref{shiftop} defined as:
\begin{equation}
    \centering
    \tilde{f}_{i, j}=f_{i+\Delta x, j+\Delta y}, \forall i, j,
    \label{conv shift}
\end{equation}
where $\Delta x$ and $\Delta y$ correspond to the horizontal and vertical shifts. Through this decomposition, we can naturally share the $1 \times 1$ convolution which implicitly included in the self-attention and the convolution. On this basis, we propose a dual-branch structure to process sequential features in parallel, named TAA, one for local feature extraction while the other for global context modeling.

Specifically, we first use three groups of standard $1 \times 1$ convolutions to linearly project and reshape the input feature sequences $E_{in}$ to $E_{feat}$, respectively. This step is equivalent to generate three sets of embeddings for self-attention (known as key, query, value). Each of $E_{feat}\in \mathbb{R}^{H_n\times T \times D_h}$, where $H_n$ is the number of multi-heads, $D_h$ represents the dimension of each head. The groups of $E_{feat}$ will serve as shared input for both branches. After subsequent operations in different modes, the outputs $E_{attn}$ and $E_{conv}$ of two branches are weighted and summed:
\begin{equation}
    \centering
    \begin{aligned}
    E_{e} = \alpha E_{attn} + (1 - \alpha) E_{conv},
    \end{aligned}
\end{equation}
where $\alpha$ is a learnable coefficient. 
Through the superposition of multi-layer architectures, our AdaPerFormer realizes the role of global-local context interaction. The details of global perception attention and local convolutional shift branch will be introduced separately below.

\paragraph{Global perception attention.}
When we use the primitive self-attention to build the transformer architecture, we notice that this dense self-attention mechanism has the time complexity of $O(n^2)$, which incurs more computational cost. Meanwhile, large-scale embedding interactions will introduce non-action information, thus affecting the accuracy of final results. Some methods \cite{DBLP:journals/corr/abs-2004-05150, DBLP:conf/nips/ZaheerGDAAOPRWY20} exploit the window attention to solve this issue in NLP task. In the TAL task, the action duration and its proportion in the video are unclear, and the fixed window cannot flexibly construct the correlation between embeddings. Without introducing extra parameters and increasing the difficulty of training, we present a simple and effective mechanism named global perception attention, shown in Fig. \ref{TAA}. The attention value of each query is calculate by following steps: 
For each $q_i$, first compute the attention score with the keys $k^w_j$ in adjacent and non-overlapping windows, where $j \in \{i-w, i, i+w\}$.
Next, we calculate the cosine similarity between $q_i$ and $q_j$ as the weight $\delta_{j}$.
Finally, we perform weighted sum over the scores and get the attention score of each $q_i$.
This window attention can be expressed as: 
\begin{equation}
\centering
\begin{aligned}
\operatorname{Attn(Q, K) = \sum_i^{T}\sum_{j}}\delta_{j} (q_i \cdot k_j^{w}),
\end{aligned}
\end{equation}
where $\delta_{j} = \mathrm{cosine}(q_i, q_j)$, and $k^w_i$ are the embeddings of a window $w$ for each $q_i$. The time complexity of our method is reduced to $O(n \times 3w)$, which decreases the computation. We use this design to ensure that each query can maximally calculate the key's attention value with a higher local correlation. 
Because of the layer-by-layer structure of the transformer, the higher-level model has wider receptive fields than the lower-level model. This mode can model a global representation that fuses all sequence information without compromising sequence completeness.
The output of the global perception attention is calculated as follows:
\begin{equation}
\centering
\begin{aligned}
head_i = \operatorname{Softmax}\left(\frac{\operatorname{Attn}(Q_i K_i^{\top})}{\sqrt{T}}\right) V_i,
\end{aligned}
\end{equation}
\begin{equation}
\centering
\begin{aligned}
E_{attn} = \operatorname{Concat}\left(head_1, head_2, \cdots, head_{H_n}\right), 
\end{aligned}
\end{equation}
where $Q_i, K_i, V_i$ are the embeddings of each head. The output $E_{attn} \in \mathbb{R}^{C_D \times T}$, will interact with the output of the local convolutional shift branch.

\begin{figure}[t]
  \centering
  \includegraphics[width=0.8\linewidth]{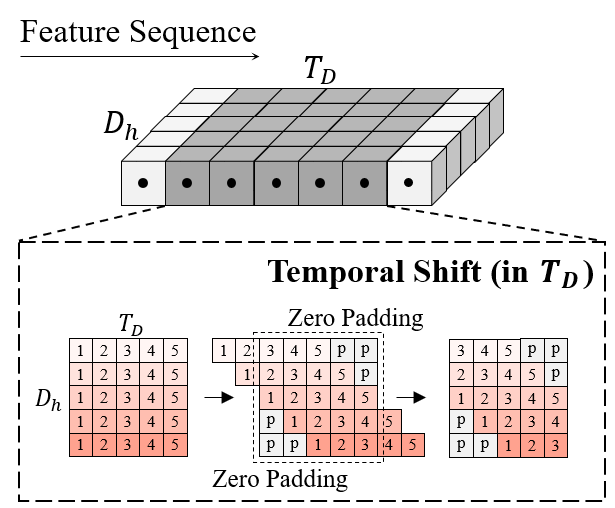}
  \caption{Overview of the temporal shift in the local convolution shift branch.}
  \label{lcs}
\end{figure}

\paragraph{Local convolutional shift.}
The primitive shift operation in Eq. \eqref{conv shift} only includes the shift within the range of the convolution kernel size, which is not applicable to long feature sequences.
Due to the continuity and diversity of actions, sequentially aggregating local contexts throughout the entire sequences can handle complex actions more flexible.
Therefore, we present a bidirectional shift to better model inter-frame and intra-frame information in every local region. 
As shown in Fig. \ref{lcs}, in a $D_h \times T_D$ feature sequence, the channel dimension $D_h$ includes the features of the temporal region, and the temporal dimension $T_D$ contains the features in each frame. The features of these two dimensions reflect the continuity and instantaneity of action, respectively. We perform shift operations in these two dimensions. Take the temporal shift in Fig. \ref{lcs} as an example. By the interlaced shift in the $D_h$ direction, each column of frame features in the $T_D$ direction interacts with the adjacent frames. The vacancies created by the shift are padded with 0. We add the shift results in both directions and form a final shift result $E_{conv}$ by residual connection. The complete process is shown in Fig. \ref{TAA}. Under the hierarchical design of the transformer decoder, all the features have fully aggregated locally. The inter-frame information is amply used to generate action boundaries, while intra-frame information helps recognize actions. 

\paragraph{Discussion}
We note the latest work ActionFormer, which achieves far ahead results on TAL. We contrast ActionFormer with our method and discuss the differences as follows:
(1) Main contribution. The main contribution of ActionFormer is to propose a single-stage anchor-free framework based on transformer. ActionFormer exhibits its powerful performance through a simple model and unique improvements to transformer encoder. However, our method follows the traditional transformer architecture and focuses on improving the attention module.
(2) Temporal-aware attention vs. local Self-attention. The original intention of our temporal-aware attention is to make the transformer have the ability to obtain local and global information at the same time. We design a dual-branch structure to to handle the local interaction of temporal context information and model the long-range global dependency. We believe that the core idea of the self-attention design is to reduce the computation while maintaining the capability of acquiring the global information. The Local self-attention used in ActionFormer proposed to limit the attention within a local window. This design is more concerned with paying attention to a certain range of temporal context within self-attention, which also achieve excellent performance.
(3) Learnable query vs. Dense prediction. Our method proposed a set of learnable query to predict actions by extracting global-local context between these queries. Follow the design of RTD-Net \cite{DBLP:conf/iccv/TanT0W21} and TadTR \cite{DBLP:journals/corr/abs-2106-10271}, we use the bipartite matching mechanism to achieve the label assignment of predictions and the ground truth. However, ActionFormer use two head to examine each moment of video sequence, and predict the action categories and the distances between current moment and the action boundaries. This dense prediction undoubtedly achieves better results. 
ActionFormer's strong performance builds a solid baseline for TAL. We hope to develop our work on this basis and jointly promote the development and progress of TAL.

\begin{table*}[t]
\centering
\resizebox{\textwidth}{!}{%
\begin{tabular}{@{}c|c|cccccc|cccc@{}}
\toprule
\multirow{2}{*}{Model} & \multirow{2}{*}{Backbone} & \multicolumn{6}{c}{THUMOS14} & \multicolumn{4}{c}{ActivityNet-1.3} \\  \cmidrule(l){3-12} 
 & & 0.3 & 0.4 & 0.5 & 0.6 & 0.7 & Avg & 0.50 & 0.75 & 0.95 & Avg \\\midrule
TAL-Net \cite{DBLP:conf/cvpr/ChaoVSRDS18} & I3D & 53.2 & 48.5 & 42.8 & 33.8 & 20.8 & 39.8 & 38.2 & 18.3 & 1.3 & 20.2 \\
PBRNet \cite{DBLP:conf/aaai/LiuW20} & I3D & 58.5 & 54.6 & 51.3 & 41.8 & 29.5 & 47.4 & 54.0 & 35.0 & 9.0 & 35.0 \\
A2Net \cite{DBLP:journals/tip/YangPZFH20} & I3D & 58.6 & 54.1 & 45.5 & 32.5 & 17.2 & 41.6 & 43.6 & 28.7 & 3.7 & 27.8 \\
ASFD \cite{DBLP:conf/cvpr/Lin0LWTWLHF21} & I3D & 67.3 & 62.4 & 55.5 & 43.7 & 31.1 & 52.0 & 53.4 & 35.3 & 6.5 & 34.4 \\
RCL \cite{wang2022rcl} & I3D & 70.1 & 62.3 & 52.9 & 42.7 & 30.7 & - & 51.7 & 35.3 & 8.0 & 34.4 \\
GTAN \cite{DBLP:conf/cvpr/LongYQTLM19} & P3D & 57.8 & 47.2 & 38.8 & - & - & - & 52.6 & 34.1 & 8.9 & 34.3 \\
VSGN \cite{DBLP:conf/iccv/ZhaoTG21} & TSN & 66.7 & 60.4 & 52.4 & 41.0 & 30.4 & 50.2 & 52.4 & 36.0 & 8.4 & 35.1 \\ 
BSN \cite{DBLP:conf/eccv/LinZSWY18} & TSN & 53.5 & 45.0 & 36.9 & 28.4 & 20.0 & 36.8 & 46.5 & 30.0 & 8.0 & 30.0 \\
BMN \cite{DBLP:conf/iccv/LinLLDW19} & TSN & 50.6 & 47.4 & 38.8 & 29.7 & 20.5 & 38.5 & 50.1 & 34.8 & 8.3 & 33.9 \\
DBG \cite{DBLP:conf/aaai/LinLWTLCWLHJ20} & TSN & 57.8 & 49.4 & 39.8 & 30.2 & 21.7 & 39.8 & - & - & - & - \\
G-TAD \cite{DBLP:conf/cvpr/XuZRTG20} & TSN & 54.5 & 47.6 & 40.3 & 30.8 & 23.4 & 39.3 & 50.4 & 34.6 & 9.0 & 34.1 \\
BC-GNN \cite{DBLP:conf/eccv/BaiWTYLL20} & TSN & 57.1 & 49.1 & 40.4 & 31.2 & 23.1 & 40.2 & 50.6 & 34.8 & 9.4 & 34.3 \\
TCANet \cite{DBLP:conf/cvpr/QingSGW0W0YGS21} & TSN & 60.6 & 53.2 & 44.6 & 36.8 & 26.7 & 44.3 & 52.3 & 6.7 & 6.9 & 35.5 \\
DCAN \cite{DBLP:conf/aaai/ChenZWL22} & TSN & 68.2 & 62.7 & 54.1 & 43.9 & 32.6 & - & 51.8 & 36.0 & 9.4 & 35.4 \\
P-GCN \cite{DBLP:conf/iccv/ZengHGTRZH19} & I3D & 63.6 & 57.8 & 49.1 & - & - & - & 48.3 & 33.2 & 3.3 & 31.1 \\
BU-TAL \cite{DBLP:conf/eccv/ZhaoXJZW020} & I3D & 53.9 & 50.7 & 45.4 & 38.0 & 28.5 & - & 43.5 & 33.9 & 9.2 & 30.1 \\
ContextLoc \cite{DBLP:conf/iccv/ZhuT00021} & I3D & 68.3 & 63.8 & 54.3 & 41.8 & 26.2 & - & \textbf{56.0} & 35.2 & 3.6 & 34.2 \\
MUSES \cite{DBLP:conf/cvpr/LiuHBDBT21} & I3D & 68.9 & 64.0 & 56.9 & 46.3 & 31.0 & 53.4 & 50.0 & 35.0 & 6.6 & 34.0 \\ 
RTD-Net \cite{DBLP:conf/iccv/TanT0W21} & I3D & 68.3 & 62.3 & 51.9 & 38.8 & 23.7 & 49.0 & 47.2 & 30.7 & 8.6 & 30.8 \\
TadTR \cite{DBLP:journals/corr/abs-2106-10271} & I3D & 74.8 & 69.1 & 60.1 & 46.6 & 32.8 & 56.7 & 52.8 & \textbf{37.1} & \textbf{10.8} & \textbf{36.1} \\
ours & I3D & 76.8  & 71.1 & 63.3 & 50.6 & 34.7 & 59.3 & 52.5 & 36.4 & 8.4 & 35.1 \\
ActionFormer \cite{DBLP:journals/corr/abs-2202-07925} & I3D & 82.1 & 77.8 & \textbf{71.0} & 59.4 & 43.9 & 66.8 & 53.5 & 36.2 & 8.2 & 35.6 \\
ActionFormer + ours & I3D & \textbf{82.3} & \textbf{78.3} & 70.7 & \textbf{59.5} & \textbf{44.8} & \textbf{67.1} & - & - & - & - \\
\bottomrule
\end{tabular}%
}
\caption{Results on the THUMOS14 and ActivityNet-1.3. We use mAP at different tIoU thresholds to report our results. The tIoU threshold is set to \{0.3, 0.4, 0.5, 0.6, 0.7\} on THUMOS14 and \{0.50, 0.75, 0.95\} on ActivityNet-1.3.}
\label{Total Result}
\end{table*}
\subsection{Training and Inference}
\paragraph{Loss function.}
Following the original design of RTD-Net \cite{DBLP:conf/iccv/TanT0W21}, we build the bipartite matching mechanism to determine the label assignment to each proposal query. We use both Distance-IoU loss \cite{DBLP:conf/aaai/ZhengWLLYR20} and $L_1$ loss in bipartite matching for complementing. The detection results are regard as $\Phi_X$ = $\{((t_s, t_e), c_n)\}^{N_X}_{n=1}$, including the distances of action boundaries $(t_s, t_e)$ and the probability of action categories $c_n$. After the label assignment of predictions and ground truth, our loss function inspired by previous work \cite{DBLP:conf/iccv/TianSCH19} and consists of two parts: the focal loss $L_{cls}$ \cite{DBLP:journals/pami/LinGGHD20} for $C$ action categories of action instances; $L_{reg}$ and $L_1$ are DIoU loss and $L_1$ loss between the normalized segment locations and the ground-truth instances, which together contribute to the boundary regression. The expression of $L_{total}$ is defined as:
\begin{equation}
\label{loss fuction}
\begin{aligned}
L_{total} = \sum^{N_X} \left[\frac{1}{N_X} L_{\mathrm{cls}} +\frac{\lambda}{N_{pos}} \mathbbm{1}_{\left\{a \neq \varnothing \right\}} \left(L_{\mathrm{reg}} + L_1 \right) \right], 
\end{aligned}
\end{equation}
where $N_{pos}$ is the total number of positive samples, $\mathbbm{1}_{\left\{a \neq \varnothing \right\}}$ is an indicator function, being 0 if the temporal segment without an action. $\lambda$ is the balance weight between the classification and regression loss. 
$L_{reg}$ and $L_1$ are only enabled when the current proposal query contains a positive sample.
\paragraph{Training details \& Inference.}
During training, we use an advanced matching scheme proposed by RTD-Net \cite{DBLP:conf/iccv/TanT0W21} to match the prediction and the ground-truth. Refer to the settings of  \cite{DBLP:conf/emnlp/LiuLGCH20}, we use AdamW \cite{DBLP:conf/iclr/LoshchilovH19} with warm-up. While in the inference stage, we generate the proposal queries from the input feature sequences and adjust accordingly in TAA.
\section{Experiments}

\subsection{Datesets and Metrics}
To verify the effectiveness of our method, we conduct extensive experiments on two widely used datasets, THUMOS14 \cite{THUMOS14} and ActivityNet-1.3 \cite{DBLP:conf/cvpr/HeilbronEGN15}. THUMOS14 consists of 413 unedited videos and 20 sports action categories. 200 of these videos are divided into the validation set, and the test set contains another 213 videos. These two sets contain 3007 and 3358 action instances, respectively. ActivityNet-1.3 is a large dataset with 19994 videos and 200 action categories. Since data for its test set is not available, we train on an annotated training set of 10,024 videos and present our results on a validation set of 4,924 videos.
Following convention, we use the standard mean average precision(mAP) at the different temporal intersections over union (tIoU) thresholds. The tIoU thresholds are set to [0.3:0.1:0.7] on THUMOS14 and [0.5:0.05:0.95] on ActivityNet-1.3. For a certain tIoU value in a given range, mAP computes the mean of the average predictions across all action categories. At the same time, we also give the average of mAP at several specific tIoUs as an overall evaluation.
\subsection{Results on THUMOS14 and ActivityNet-1.3}
\paragraph{Implementation details.}
Our model is implemented in Pytorch. All the experiments are conducted on NVIDIA Tesla A100 GPU. We use I3D \cite{DBLP:conf/cvpr/CarreiraZ17} pretrained on Kinetics to extract the feature sequences. Taking THUMOS14 as an example, we extract 1024-D features before the last fully connected layer of I3D, and concatenate them to 2048-D as the input of our model, following the same method as the previous work. We trained the AdaPerFormer for 30 epochs in THUMOS14 and 20 epochs in ActivityNet-1.3 with a linear warm-up of 5 epochs. We used AdamW by initial learning rate $10^{-4}, 10^{-3}$ respectively with a cosine learning rate decay, and the weight decay was set to $10^{-4}$. The batch size was set to 8 and 16, respectively, and the weight $\lambda$ of $L_{reg}$ was set to 1. For the hyper-parameters in TAA, the attention window size was set to 5, and the shift size was set to 9 and 7, respectively. We compare our method to a set of baselines. For fair comparisons, we try our best to tune the parameters for the competitors, and choose their best possible results to compare.

\paragraph{Results.}
Table \ref{Total Result} shows the total results. For the THUMOS14, We compare our method with classical and recent approaches in the TAL. Our method achieves an average mAP of 59.3\% with tIoU setting in [0.3:0.1:0.7]. Not only does our method achieves significant improvement over previous anchor-based and boundary-based methods, but it is worth noting that our method has obvious advantages on all metrics compared to the query-based methods RTD-Net and TadTR. The state-of-the-art method, ActionFormer, achieves extraordinary performance on this dataset. For comparison, we use the settings of ActionFormer, and replace the local self-attention in the encoder with our TAA. We achieved a slight improvement of 0.3\% on 
ActionFormer's excellent result, which also reflects the superiority of the ActionFormer. For the ActivityNet-1.3, AdaPerFormer achieves an average mAP of 35.1\% and demonstrates competitive results in this challenging dataset. 
\subsection{Ablation Study}
We conduct extensive ablation experiments on the THUMOS14 dataset to verify our proposed method's performance and the model design's rationality. All results are based on I3D features and keep the same random seed for training.We report mAP at tIoU=0.3, 0.5 and 0.7, the average mAP in [0.3:0.1:0.7] for all results. Unless otherwise specified, all results, except for experimental variables, use the settings with the best results.
\paragraph{Model design.}
We explore the settings of different model branches. The dense self-attention is used to build our baseline upon the transformer architecture. Our global perception attention has a further improvement on this basis, especially when the tIoU=0.5 and 0.7. This validates our unique window attention design. Furthermore, the dual-branch design also leads to a more significant performance improvement, which further demonstrates the effectiveness of our local convolutional shift branch.
\begin{table}[t]
\centering
% \resizebox{\columnwidth}{!}{%
\begin{tabular}{ccc|cccc}
\toprule
DSA & GPA & LCS & 0.3 & 0.5 & 0.7 & Avg \\ \midrule
$\checkmark$ &  &  & 74.2 & 58.9 & 30.8 & 55.5 \\ 
 & $\checkmark$ &  & 75.5 & 62.0 & 33.1 & 57.9 \\ 
$\checkmark$ &  & $\checkmark$ & 74.9 & 61.6 & 33.4 & 57.6 \\ 
 & $\checkmark$ & $\checkmark$ & \textbf{76.8} & \textbf{63.3} & \textbf{34.7} & \textbf{59.3} \\ \bottomrule
\end{tabular}%
% }
\caption{Ablation study on different configurations. We choose the dense self-attention (DSA) as the base transformer, and evaluate the effect of the global perception attention (GPA) and local convolutional shift (LCS) branches.}
\label{1}
\end{table}
\paragraph{Global perception attention.}
We verified the effect of window size in our global perception attention branch. Table \ref{window} shows results. The window size in table represent the size of key, so each query can interact with $3w$ of keys. Note that the dense attention will bring significant performance degradation. This phenomenon further illustrates that dense self-attention will bring more unnecessary embedding interactions and a high computational cost. However, our method can model the global interaction of feature sequences without incurring extra computation.
\begin{table}[t]
\centering
% \resizebox{.9\columnwidth}{!}{%
\begin{tabular}{c|cccc}
\toprule
Win. size & 0.3 & 0.5 & 0.7 & Avg\\ \midrule
3 & 76.4 & 62.5 & 34.1 & 58.8 \\ 
5 & \textbf{76.8} & \textbf{63.3} & \textbf{34.7} & \textbf{59.3} \\
7 & 76.3 & 62.0 & 34.6 & 59.0 \\ 
all & 74.9 & 61.6 & 33.4 & 57.6 \\ \bottomrule
\end{tabular}%
% }
\caption{Ablation study on window size in global perception attention branch. `all' represent the dense self-attention.}
\label{window}
\end{table}
\paragraph{Local convolutional shift.}
Table \ref{shift setting} shows the results of using different shift operations in different parts of the encoder and decoder. From the results, the bidirectional shift has a certain improvement compared with the general shift. Note that using a bidirectional shift in the encoder part and a general shift in the encoder part can make the model achieve the best result.
\begin{table}[t]
\centering
% \resizebox{\columnwidth}{!}{%
\begin{tabular}{cc|cccc}
\toprule
Encoder & Decoder & 0.3 & 0.5 & 0.7 & Avg \\ \midrule
GS & GS & 75.6 & 52.0 & 33.7 & 58.0 \\
GS & BS & 74.6 & 61.2 & \textbf{36.6} & 58.3 \\
BS & GS & 76.8 & \textbf{63.3} & 34.7 & \textbf{59.3} \\
BS & BS & \textbf{77.1} & 62.4 & 35.0 & 59.1 \\ \bottomrule
\end{tabular}%
% }
\caption{Ablation study on the local convolutional shift setting. At different stages, we explore the effects of two shift modes, general shift (GS) and bidirectional shift (BS).}
\label{shift setting}
\end{table}
\paragraph{TAA weight.}
When weighing the two branches of TAA, Since the weights required in different layers may be different, we try several strategies of the weight setting. Table \ref{weight init} reports the results. As can be seen, by setting learnable parameters in each layer of the network and fixing their weight sum to 1, the network can learn more suitable parameters with better performance. Note that using two learnable coefficients increases the difficulty of training, and also results in lower performance under the same conditions.
\begin{table}[t]
\centering
% \resizebox{\columnwidth}{!}{%
\begin{tabular}{cc|cccc}
\toprule
$\alpha_1$ & $\alpha_2$ & 0.3 & 0.5 & 0.7 & Avg \\ \midrule
1 & 1 & 76.3 & 63.1 & 34.2 & 58.9 \\
1 & $\alpha$ & 76.5 & 63.0 & \textbf{34.7} & 59.0 \\
$\alpha$ & 1 & \textbf{77.0} & \textbf{63.4} & 34.5 & 59.2 \\
$\alpha$ & 1-$\alpha$ & 76.8 & 63.3 & \textbf{34.7} & \textbf{59.3} \\
$\alpha_1$ & $\alpha_2$ & 75.9 & 62.3 & 34.3 & 58.3 \\ \bottomrule
\end{tabular}%
% }
\caption{Ablation study on the TAA weight setting.}
\vspace{8pt}
\label{weight init}
% \end{table}

% \begin{table}[t]
\centering
% \resizebox{\columnwidth}{!}{%
\begin{tabular}{c|cccc}
\toprule
shift size & 0.3 & 0.5 & 0.7 & Avg \\ \midrule
5 & 75.9 & 62.6 & 34.5 & 58.5 \\
7 & \textbf{77.1} & 62.3 & 33.4 & 59.0 \\
9 & 76.8 & \textbf{63.3} & 34.7 & \textbf{59.3} \\
11 & 76.4 & 62.8 & \textbf{34.8} & 58.9 \\ \bottomrule
\end{tabular}%
% }
\caption{Ablation study on shift size.}
\label{size}
\end{table}

\paragraph{Shift size.}
We conduct ablation studies on the size of the local convolution shift operation. Table  \ref{size} shows all the above results. In our experiments, we select the shift size in [5, 7, 9, 11] for horizontal comparison. We notice that an excessively large shift size will lead to a slight decrease in performance, since an excessively large shift distance will destroy the spatial information of the sequence.

\section{Concluding Remarks}

\label{conclusion}
In this paper, we have presented an end-to-end framework named AdaPerFormer for TAL. AdaPerFormer can efficiently process the entire feature sequences and model local-global dependencies by dual-branch TAA mechanism, thereby directly generate action categories and boundaries end-to-end without extra processing.  Extensive experiments on the THUMOS14 and ActivityNet-1.3 datasets demonstrate the advance of our method in TAL.
However, most, if not all, of current methods use pre-trained video features for TAL, which is also the limitation of our method. In addition, training with human-annotated video data makes our method not transferable to real-world scenarios well, which is also a big challenge in this field. In the future, we expect to make full use of video information in the pre-training stage to build a more general framework for action recognition and temporal action localization.

\nocite{*}
\bibliography{aaai23}

\begin{thebibliography}{65}
\providecommand{\natexlab}[1]{#1}

\bibitem[{Alwassel et~al.(2018)Alwassel, Caba~Heilbron, Escorcia, and
  Ghanem}]{alwassel_2018_detad}
Alwassel, H.; Caba~Heilbron, F.; Escorcia, V.; and Ghanem, B. 2018.
\newblock Diagnosing Error in Temporal Action Detectors.
\newblock In \emph{ECCV}.

\bibitem[{Alwassel, Giancola, and Ghanem(2021)}]{DBLP:conf/iccvw/AlwasselGG21}
Alwassel, H.; Giancola, S.; and Ghanem, B. 2021.
\newblock {TSP:} Temporally-Sensitive Pretraining of Video Encoders for
  Localization Tasks.
\newblock In \emph{ICCV Workshops}, 3166--3176.

\bibitem[{Bai et~al.(2020)Bai, Wang, Tong, Yang, Liu, and
  Liu}]{DBLP:conf/eccv/BaiWTYLL20}
Bai, Y.; Wang, Y.; Tong, Y.; Yang, Y.; Liu, Q.; and Liu, J. 2020.
\newblock Boundary Content Graph Neural Network for Temporal Action Proposal
  Generation.
\newblock In \emph{ECCV}, 121--137.

\bibitem[{Beltagy, Peters, and Cohan(2020)}]{DBLP:journals/corr/abs-2004-05150}
Beltagy, I.; Peters, M.~E.; and Cohan, A. 2020.
\newblock Longformer: The Long-Document Transformer.
\newblock \emph{CoRR}, abs/2004.05150.

\bibitem[{Buch et~al.(2017{\natexlab{a}})Buch, Escorcia, Ghanem, Fei{-}Fei, and
  Niebles}]{DBLP:conf/bmvc/BuchEGFN17}
Buch, S.; Escorcia, V.; Ghanem, B.; Fei{-}Fei, L.; and Niebles, J.~C.
  2017{\natexlab{a}}.
\newblock End-to-End, Single-Stream Temporal Action Detection in Untrimmed
  Videos.
\newblock In \emph{BMVC}.

\bibitem[{Buch et~al.(2017{\natexlab{b}})Buch, Escorcia, Shen, Ghanem, and
  Niebles}]{DBLP:conf/cvpr/BuchESGN17}
Buch, S.; Escorcia, V.; Shen, C.; Ghanem, B.; and Niebles, J.~C.
  2017{\natexlab{b}}.
\newblock {SST:} Single-Stream Temporal Action Proposals.
\newblock In \emph{CVPR}, 6373--6382.

\bibitem[{Carreira and Zisserman(2017)}]{DBLP:conf/cvpr/CarreiraZ17}
Carreira, J.; and Zisserman, A. 2017.
\newblock Quo Vadis, Action Recognition? {A} New Model and the Kinetics
  Dataset.
\newblock In \emph{CVPR}, 4724--4733.

\bibitem[{Chao et~al.(2018)Chao, Vijayanarasimhan, Seybold, Ross, Deng, and
  Sukthankar}]{DBLP:conf/cvpr/ChaoVSRDS18}
Chao, Y.; Vijayanarasimhan, S.; Seybold, B.; Ross, D.~A.; Deng, J.; and
  Sukthankar, R. 2018.
\newblock Rethinking the Faster {R-CNN} Architecture for Temporal Action
  Localization.
\newblock In \emph{CVPR}, 1130--1139.

\bibitem[{Chen et~al.(2022)Chen, Zheng, Wang, and
  Lu}]{DBLP:conf/aaai/ChenZWL22}
Chen, G.; Zheng, Y.; Wang, L.; and Lu, T. 2022.
\newblock {DCAN:} Improving Temporal Action Detection via Dual Context
  Aggregation.
\newblock In \emph{AAAI}, 248--257.

\bibitem[{Dosovitskiy et~al.(2021)Dosovitskiy, Beyer, Kolesnikov, Weissenborn,
  Zhai, Unterthiner, Dehghani, Minderer, Heigold, Gelly, Uszkoreit, and
  Houlsby}]{DBLP:conf/iclr/DosovitskiyB0WZ21}
Dosovitskiy, A.; Beyer, L.; Kolesnikov, A.; Weissenborn, D.; Zhai, X.;
  Unterthiner, T.; Dehghani, M.; Minderer, M.; Heigold, G.; Gelly, S.;
  Uszkoreit, J.; and Houlsby, N. 2021.
\newblock An Image is Worth 16x16 Words: Transformers for Image Recognition at
  Scale.
\newblock In \emph{ICLR}.

\bibitem[{Escorcia et~al.(2016)Escorcia, Heilbron, Niebles, and
  Ghanem}]{DBLP:conf/eccv/EscorciaHNG16}
Escorcia, V.; Heilbron, F.~C.; Niebles, J.~C.; and Ghanem, B. 2016.
\newblock DAPs: Deep Action Proposals for Action Understanding.
\newblock In \emph{ECCV}, 768--784.

\bibitem[{Feichtenhofer, Pinz, and
  Wildes(2016)}]{DBLP:conf/nips/FeichtenhoferPW16}
Feichtenhofer, C.; Pinz, A.; and Wildes, R.~P. 2016.
\newblock Spatiotemporal Residual Networks for Video Action Recognition.
\newblock In \emph{NeurIPS}, 3468--3476.

\bibitem[{Gao, Yang, and Nevatia(2017)}]{DBLP:conf/bmvc/GaoYN17}
Gao, J.; Yang, Z.; and Nevatia, R. 2017.
\newblock Cascaded Boundary Regression for Temporal Action Detection.
\newblock In \emph{BMVC}.

\bibitem[{Gao et~al.(2017)Gao, Yang, Sun, Chen, and
  Nevatia}]{DBLP:conf/iccv/GaoYSCN17}
Gao, J.; Yang, Z.; Sun, C.; Chen, K.; and Nevatia, R. 2017.
\newblock {TURN} {TAP:} Temporal Unit Regression Network for Temporal Action
  Proposals.
\newblock In \emph{ICCV}, 3648--3656.

\bibitem[{Heilbron et~al.(2015)Heilbron, Escorcia, Ghanem, and
  Niebles}]{DBLP:conf/cvpr/HeilbronEGN15}
Heilbron, F.~C.; Escorcia, V.; Ghanem, B.; and Niebles, J.~C. 2015.
\newblock ActivityNet: {A} large-scale video benchmark for human activity
  understanding.
\newblock In \emph{CVPR}, 961--970.

\bibitem[{Heilbron, Niebles, and Ghanem(2016)}]{DBLP:conf/cvpr/HeilbronNG16}
Heilbron, F.~C.; Niebles, J.~C.; and Ghanem, B. 2016.
\newblock Fast Temporal Activity Proposals for Efficient Detection of Human
  Actions in Untrimmed Videos.
\newblock In \emph{CVPR}, 1914--1923.

\bibitem[{Hu et~al.(2018)Hu, Shen, Albanie, Sun, and
  Vedaldi}]{DBLP:conf/nips/HuSASV18}
Hu, J.; Shen, L.; Albanie, S.; Sun, G.; and Vedaldi, A. 2018.
\newblock Gather-Excite: Exploiting Feature Context in Convolutional Neural
  Networks.
\newblock In \emph{NeurIPS}, 9423--9433.

\bibitem[{Hu et~al.(2020)Hu, Shen, Albanie, Sun, and
  Wu}]{DBLP:journals/pami/HuSASW20}
Hu, J.; Shen, L.; Albanie, S.; Sun, G.; and Wu, E. 2020.
\newblock Squeeze-and-Excitation Networks.
\newblock \emph{TPAMI}, 42(8): 2011--2023.

\bibitem[{Hu, Shen, and Sun(2018)}]{DBLP:conf/cvpr/HuSS18}
Hu, J.; Shen, L.; and Sun, G. 2018.
\newblock Squeeze-and-Excitation Networks.
\newblock In \emph{CVPR}, 7132--7141.

\bibitem[{Idrees et~al.(2017)Idrees, Zamir, Jiang, Gorban, Laptev, Sukthankar,
  and Shah}]{DBLP:journals/cviu/IdreesZJGLSS17}
Idrees, H.; Zamir, A.~R.; Jiang, Y.; Gorban, A.; Laptev, I.; Sukthankar, R.;
  and Shah, M. 2017.
\newblock The {THUMOS} challenge on action recognition for videos `in the
  wild".
\newblock \emph{CVIU}, 155: 1--23.

\bibitem[{Jiang et~al.(2014)Jiang, Liu, Roshan~Zamir, Toderici, Laptev, Shah,
  and Sukthankar}]{THUMOS14}
Jiang, Y.-G.; Liu, J.; Roshan~Zamir, A.; Toderici, G.; Laptev, I.; Shah, M.;
  and Sukthankar, R. 2014.
\newblock {THUMOS} Challenge: Action Recognition with a Large Number of
  Classes.
\newblock \url{http://crcv.ucf.edu/THUMOS14/}.

\bibitem[{Li et~al.(2019)Li, Jin, Xuan, Zhou, Chen, Wang, and
  Yan}]{DBLP:conf/nips/LiJXZCWY19}
Li, S.; Jin, X.; Xuan, Y.; Zhou, X.; Chen, W.; Wang, Y.; and Yan, X. 2019.
\newblock Enhancing the Locality and Breaking the Memory Bottleneck of
  Transformer on Time Series Forecasting.
\newblock In \emph{NeurIPS}, 5244--5254.

\bibitem[{Lin et~al.(2020{\natexlab{a}})Lin, Li, Wang, Tai, Luo, Cui, Wang, Li,
  Huang, and Ji}]{DBLP:conf/aaai/LinLWTLCWLHJ20}
Lin, C.; Li, J.; Wang, Y.; Tai, Y.; Luo, D.; Cui, Z.; Wang, C.; Li, J.; Huang,
  F.; and Ji, R. 2020{\natexlab{a}}.
\newblock Fast Learning of Temporal Action Proposal via Dense Boundary
  Generator.
\newblock In \emph{AAAI}, 11499--11506.

\bibitem[{Lin et~al.(2021)Lin, Xu, Luo, Wang, Tai, Wang, Li, Huang, and
  Fu}]{DBLP:conf/cvpr/Lin0LWTWLHF21}
Lin, C.; Xu, C.; Luo, D.; Wang, Y.; Tai, Y.; Wang, C.; Li, J.; Huang, F.; and
  Fu, Y. 2021.
\newblock Learning Salient Boundary Feature for Anchor-free Temporal Action
  Localization.
\newblock In \emph{CVPR}, 3320--3329.

\bibitem[{Lin et~al.(2020{\natexlab{b}})Lin, Goyal, Girshick, He, and
  Doll{\'{a}}r}]{DBLP:journals/pami/LinGGHD20}
Lin, T.; Goyal, P.; Girshick, R.~B.; He, K.; and Doll{\'{a}}r, P.
  2020{\natexlab{b}}.
\newblock Focal Loss for Dense Object Detection.
\newblock \emph{TPAMI}, 42(2): 318--327.

\bibitem[{Lin et~al.(2019)Lin, Liu, Li, Ding, and
  Wen}]{DBLP:conf/iccv/LinLLDW19}
Lin, T.; Liu, X.; Li, X.; Ding, E.; and Wen, S. 2019.
\newblock {BMN:} Boundary-Matching Network for Temporal Action Proposal
  Generation.
\newblock In \emph{ICCV}, 3888--3897.

\bibitem[{Lin, Zhao, and Shou(2017)}]{DBLP:conf/mm/LinZS17}
Lin, T.; Zhao, X.; and Shou, Z. 2017.
\newblock Single Shot Temporal Action Detection.
\newblock In \emph{ACMMM}, 988--996.

\bibitem[{Lin et~al.(2018)Lin, Zhao, Su, Wang, and
  Yang}]{DBLP:conf/eccv/LinZSWY18}
Lin, T.; Zhao, X.; Su, H.; Wang, C.; and Yang, M. 2018.
\newblock {BSN:} Boundary Sensitive Network for Temporal Action Proposal
  Generation.
\newblock In \emph{ECCV}, 3--21.

\bibitem[{Liu et~al.(2020)Liu, Liu, Gao, Chen, and
  Han}]{DBLP:conf/emnlp/LiuLGCH20}
Liu, L.; Liu, X.; Gao, J.; Chen, W.; and Han, J. 2020.
\newblock Understanding the Difficulty of Training Transformers.
\newblock In \emph{EMNLP}, 5747--5763.

\bibitem[{Liu and Wang(2020)}]{DBLP:conf/aaai/LiuW20}
Liu, Q.; and Wang, Z. 2020.
\newblock Progressive Boundary Refinement Network for Temporal Action
  Detection.
\newblock In \emph{AAAI}, 11612--11619.

\bibitem[{Liu et~al.(2021{\natexlab{a}})Liu, Hu, Bai, Ding, Bai, and
  Torr}]{DBLP:conf/cvpr/LiuHBDBT21}
Liu, X.; Hu, Y.; Bai, S.; Ding, F.; Bai, X.; and Torr, P. H.~S.
  2021{\natexlab{a}}.
\newblock Multi-Shot Temporal Event Localization: {A} Benchmark.
\newblock In \emph{CVPR}, 12596--12606.

\bibitem[{Liu et~al.(2021{\natexlab{b}})Liu, Wang, Hu, Tang, Bai, and
  Bai}]{DBLP:journals/corr/abs-2106-10271}
Liu, X.; Wang, Q.; Hu, Y.; Tang, X.; Bai, S.; and Bai, X. 2021{\natexlab{b}}.
\newblock End-to-end Temporal Action Detection with Transformer.
\newblock \emph{CoRR}, abs/2106.10271.

\bibitem[{Liu et~al.(2019)Liu, Ma, Zhang, Liu, and
  Chang}]{DBLP:conf/cvpr/Liu0Z0C19}
Liu, Y.; Ma, L.; Zhang, Y.; Liu, W.; and Chang, S. 2019.
\newblock Multi-Granularity Generator for Temporal Action Proposal.
\newblock In \emph{CVPR}, 3604--3613.

\bibitem[{Liu et~al.(2021{\natexlab{c}})Liu, Lin, Cao, Hu, Wei, Zhang, Lin, and
  Guo}]{DBLP:conf/iccv/LiuL00W0LG21}
Liu, Z.; Lin, Y.; Cao, Y.; Hu, H.; Wei, Y.; Zhang, Z.; Lin, S.; and Guo, B.
  2021{\natexlab{c}}.
\newblock Swin Transformer: Hierarchical Vision Transformer using Shifted
  Windows.
\newblock In \emph{ICCV}, 9992--10002.

\bibitem[{Long et~al.(2019)Long, Yao, Qiu, Tian, Luo, and
  Mei}]{DBLP:conf/cvpr/LongYQTLM19}
Long, F.; Yao, T.; Qiu, Z.; Tian, X.; Luo, J.; and Mei, T. 2019.
\newblock Gaussian Temporal Awareness Networks for Action Localization.
\newblock In \emph{CVPR}, 344--353.

\bibitem[{Loshchilov and Hutter(2019)}]{DBLP:conf/iclr/LoshchilovH19}
Loshchilov, I.; and Hutter, F. 2019.
\newblock Decoupled Weight Decay Regularization.
\newblock In \emph{ICLR}.

\bibitem[{Pan et~al.(2016)Pan, Xu, Yang, Wu, and
  Zhuang}]{DBLP:conf/cvpr/PanXYWZ16}
Pan, P.; Xu, Z.; Yang, Y.; Wu, F.; and Zhuang, Y. 2016.
\newblock Hierarchical Recurrent Neural Encoder for Video Representation with
  Application to Captioning.
\newblock In \emph{CVPR}, 1029--1038.

\bibitem[{Pan et~al.(2021)Pan, Ge, Lu, Song, Chen, Huang, and
  Huang}]{DBLP:journals/corr/abs-2111-14556}
Pan, X.; Ge, C.; Lu, R.; Song, S.; Chen, G.; Huang, Z.; and Huang, G. 2021.
\newblock On the Integration of Self-Attention and Convolution.
\newblock \emph{CoRR}, abs/2111.14556.

\bibitem[{Peng et~al.(2021)Peng, Huang, Gu, Xie, Wang, Jiao, and
  Ye}]{DBLP:conf/iccv/PengHGXWJY21}
Peng, Z.; Huang, W.; Gu, S.; Xie, L.; Wang, Y.; Jiao, J.; and Ye, Q. 2021.
\newblock Conformer: Local Features Coupling Global Representations for Visual
  Recognition.
\newblock In \emph{ICCV}, 357--366.

\bibitem[{Qing et~al.(2021)Qing, Su, Gan, Wang, Wu, Wang, Qiao, Yan, Gao, and
  Sang}]{DBLP:conf/cvpr/QingSGW0W0YGS21}
Qing, Z.; Su, H.; Gan, W.; Wang, D.; Wu, W.; Wang, X.; Qiao, Y.; Yan, J.; Gao,
  C.; and Sang, N. 2021.
\newblock Temporal Context Aggregation Network for Temporal Action Proposal
  Refinement.
\newblock In \emph{CVPR}, 485--494.

\bibitem[{Qiu, Yao, and Mei(2017)}]{DBLP:conf/iccv/QiuYM17}
Qiu, Z.; Yao, T.; and Mei, T. 2017.
\newblock Learning Spatio-Temporal Representation with Pseudo-3D Residual
  Networks.
\newblock In \emph{ICCV}, 5534--5542.

\bibitem[{Simonyan and Zisserman(2014)}]{DBLP:conf/nips/SimonyanZ14}
Simonyan, K.; and Zisserman, A. 2014.
\newblock Two-Stream Convolutional Networks for Action Recognition in Videos.
\newblock In \emph{NeurIPS}, 568--576.

\bibitem[{Tan et~al.(2021)Tan, Tang, Wang, and Wu}]{DBLP:conf/iccv/TanT0W21}
Tan, J.; Tang, J.; Wang, L.; and Wu, G. 2021.
\newblock Relaxed Transformer Decoders for Direct Action Proposal Generation.
\newblock In \emph{ICCV}, 13506--13515.

\bibitem[{Tian et~al.(2019)Tian, Shen, Chen, and He}]{DBLP:conf/iccv/TianSCH19}
Tian, Z.; Shen, C.; Chen, H.; and He, T. 2019.
\newblock {FCOS:} Fully Convolutional One-Stage Object Detection.
\newblock In \emph{ICCV}, 9626--9635.

\bibitem[{Tran et~al.(2018)Tran, Wang, Torresani, Ray, LeCun, and
  Paluri}]{DBLP:conf/cvpr/TranWTRLP18}
Tran, D.; Wang, H.; Torresani, L.; Ray, J.; LeCun, Y.; and Paluri, M. 2018.
\newblock A Closer Look at Spatiotemporal Convolutions for Action Recognition.
\newblock In \emph{CVPR}, 6450--6459.

\bibitem[{Venugopalan et~al.(2015)Venugopalan, Rohrbach, Donahue, Mooney,
  Darrell, and Saenko}]{DBLP:conf/iccv/VenugopalanRDMD15}
Venugopalan, S.; Rohrbach, M.; Donahue, J.; Mooney, R.~J.; Darrell, T.; and
  Saenko, K. 2015.
\newblock Sequence to Sequence - Video to Text.
\newblock In \emph{ICCV}, 4534--4542.

\bibitem[{Wang et~al.(2016)Wang, Xiong, Wang, Qiao, Lin, Tang, and
  Gool}]{DBLP:conf/eccv/WangXW0LTG16}
Wang, L.; Xiong, Y.; Wang, Z.; Qiao, Y.; Lin, D.; Tang, X.; and Gool, L.~V.
  2016.
\newblock Temporal Segment Networks: Towards Good Practices for Deep Action
  Recognition.
\newblock In \emph{ECCV}, 20--36.

\bibitem[{Wang et~al.(2022)Wang, Zhang, Zheng, and Pan}]{wang2022rcl}
Wang, Q.; Zhang, Y.; Zheng, Y.; and Pan, P. 2022.
\newblock RCL: Recurrent Continuous Localization for Temporal Action Detection.
\newblock In \emph{CVPR}, 13566--13575.

\bibitem[{Wu et~al.(2021)Wu, Xiao, Codella, Liu, Dai, Yuan, and
  Zhang}]{DBLP:conf/iccv/WuXCLDY021}
Wu, H.; Xiao, B.; Codella, N.; Liu, M.; Dai, X.; Yuan, L.; and Zhang, L. 2021.
\newblock CvT: Introducing Convolutions to Vision Transformers.
\newblock In \emph{ICCV}, 22--31.

\bibitem[{Xiao et~al.(2021)Xiao, Singh, Mintun, Darrell, Doll{\'{a}}r, and
  Girshick}]{DBLP:conf/nips/XiaoSMDDG21}
Xiao, T.; Singh, M.; Mintun, E.; Darrell, T.; Doll{\'{a}}r, P.; and Girshick,
  R.~B. 2021.
\newblock Early Convolutions Help Transformers See Better.
\newblock In \emph{NeurIPS}, 30392--30400.

\bibitem[{Xie et~al.(2021)Xie, Wang, Yu, Anandkumar, Alvarez, and
  Luo}]{DBLP:conf/nips/XieWYAAL21}
Xie, E.; Wang, W.; Yu, Z.; Anandkumar, A.; Alvarez, J.~M.; and Luo, P. 2021.
\newblock SegFormer: Simple and Efficient Design for Semantic Segmentation with
  Transformers.
\newblock In \emph{NeurIPS}, 12077--12090.

\bibitem[{Xu, Das, and Saenko(2017)}]{DBLP:conf/iccv/XuDS17}
Xu, H.; Das, A.; and Saenko, K. 2017.
\newblock {R-C3D:} Region Convolutional 3D Network for Temporal Activity
  Detection.
\newblock In \emph{ICCV}, 5794--5803.

\bibitem[{Xu et~al.(2020)Xu, Zhao, Rojas, Thabet, and
  Ghanem}]{DBLP:conf/cvpr/XuZRTG20}
Xu, M.; Zhao, C.; Rojas, D.~S.; Thabet, A.~K.; and Ghanem, B. 2020.
\newblock {G-TAD:} Sub-Graph Localization for Temporal Action Detection.
\newblock In \emph{CVPR}, 10153--10162.

\bibitem[{Yang et~al.(2020)Yang, Peng, Zhang, Fu, and
  Han}]{DBLP:journals/tip/YangPZFH20}
Yang, L.; Peng, H.; Zhang, D.; Fu, J.; and Han, J. 2020.
\newblock Revisiting Anchor Mechanisms for Temporal Action Localization.
\newblock \emph{TIP}, 29: 8535--8548.

\bibitem[{Zaheer et~al.(2020)Zaheer, Guruganesh, Dubey, Ainslie, Alberti,
  Onta{\~{n}}{\'{o}}n, Pham, Ravula, Wang, Yang, and
  Ahmed}]{DBLP:conf/nips/ZaheerGDAAOPRWY20}
Zaheer, M.; Guruganesh, G.; Dubey, K.~A.; Ainslie, J.; Alberti, C.;
  Onta{\~{n}}{\'{o}}n, S.; Pham, P.; Ravula, A.; Wang, Q.; Yang, L.; and Ahmed,
  A. 2020.
\newblock Big Bird: Transformers for Longer Sequences.
\newblock In \emph{NeurIPS}, 17283--17297.

\bibitem[{Zeng et~al.(2019)Zeng, Huang, Gan, Tan, Rong, Zhao, and
  Huang}]{DBLP:conf/iccv/ZengHGTRZH19}
Zeng, R.; Huang, W.; Gan, C.; Tan, M.; Rong, Y.; Zhao, P.; and Huang, J. 2019.
\newblock Graph Convolutional Networks for Temporal Action Localization.
\newblock In \emph{ICCV}, 7093--7102.

\bibitem[{Zeng et~al.(2020)Zeng, Xu, Huang, Chen, Tan, and
  Gan}]{DBLP:conf/cvpr/ZengXHCTG20}
Zeng, R.; Xu, H.; Huang, W.; Chen, P.; Tan, M.; and Gan, C. 2020.
\newblock Dense Regression Network for Video Grounding.
\newblock In \emph{CVPR}, 10284--10293.

\bibitem[{Zhang, Wu, and Li(2022)}]{DBLP:journals/corr/abs-2202-07925}
Zhang, C.; Wu, J.; and Li, Y. 2022.
\newblock ActionFormer: Localizing Moments of Actions with Transformers.
\newblock \emph{CoRR}, abs/2202.07925.

\bibitem[{Zhang et~al.(2020)Zhang, Zhao, Zhao, Wang, Liu, and
  Gao}]{DBLP:conf/cvpr/ZhangZZWLG20}
Zhang, Z.; Zhao, Z.; Zhao, Y.; Wang, Q.; Liu, H.; and Gao, L. 2020.
\newblock Where Does It Exist: Spatio-Temporal Video Grounding for Multi-Form
  Sentences.
\newblock In \emph{CVPR}, 10665--10674.

\bibitem[{Zhao, Thabet, and Ghanem(2021)}]{DBLP:conf/iccv/ZhaoTG21}
Zhao, C.; Thabet, A.~K.; and Ghanem, B. 2021.
\newblock Video Self-Stitching Graph Network for Temporal Action Localization.
\newblock In \emph{ICCV}, 13638--13647.

\bibitem[{Zhao et~al.(2020)Zhao, Xie, Ju, Zhang, Wang, and
  Tian}]{DBLP:conf/eccv/ZhaoXJZW020}
Zhao, P.; Xie, L.; Ju, C.; Zhang, Y.; Wang, Y.; and Tian, Q. 2020.
\newblock Bottom-Up Temporal Action Localization with Mutual Regularization.
\newblock In \emph{ECCV}, 539--555.

\bibitem[{Zhao et~al.(2017)Zhao, Xiong, Wang, Wu, Tang, and
  Lin}]{DBLP:conf/iccv/ZhaoXWWTL17}
Zhao, Y.; Xiong, Y.; Wang, L.; Wu, Z.; Tang, X.; and Lin, D. 2017.
\newblock Temporal Action Detection with Structured Segment Networks.
\newblock In \emph{ICCV}, 2933--2942.

\bibitem[{Zheng et~al.(2020)Zheng, Wang, Liu, Li, Ye, and
  Ren}]{DBLP:conf/aaai/ZhengWLLYR20}
Zheng, Z.; Wang, P.; Liu, W.; Li, J.; Ye, R.; and Ren, D. 2020.
\newblock Distance-IoU Loss: Faster and Better Learning for Bounding Box
  Regression.
\newblock In \emph{AAAI}, 12993--13000.

\bibitem[{Zhu et~al.(2021{\natexlab{a}})Zhu, Su, Lu, Li, Wang, and
  Dai}]{DBLP:conf/iclr/ZhuSLLWD21}
Zhu, X.; Su, W.; Lu, L.; Li, B.; Wang, X.; and Dai, J. 2021{\natexlab{a}}.
\newblock Deformable {DETR:} Deformable Transformers for End-to-End Object
  Detection.
\newblock In \emph{ICLR}.

\bibitem[{Zhu et~al.(2021{\natexlab{b}})Zhu, Tang, Wang, Zheng, and
  Hua}]{DBLP:conf/iccv/ZhuT00021}
Zhu, Z.; Tang, W.; Wang, L.; Zheng, N.; and Hua, G. 2021{\natexlab{b}}.
\newblock Enriching Local and Global Contexts for Temporal Action Localization.
\newblock In \emph{ICCV}, 13496--13505.

\end{thebibliography}

\end{document}